\documentclass[runningheads]{llncs}
\usepackage{graphicx}

\usepackage{tikz}
\usepackage{comment}
\usepackage{amsmath,amssymb} 
\usepackage{xcolor}
\usepackage{booktabs}
\usepackage{multirow}
\usepackage{times}
\usepackage{epsfig}
\usepackage{graphicx}
\usepackage{amssymb}
\usepackage{adjustbox}
\usepackage{xcolor}
\usepackage{textcomp}
\usepackage[pagebackref=true,breaklinks=true,colorlinks,urlcolor=darkgray,bookmarks=false]{hyperref}
\usepackage{calc}
\usepackage{tabularx}
\usepackage{booktabs}
\usepackage{ragged2e}
\usepackage{xcolor}
\usepackage{commath}
\usepackage{ulem}

\usepackage{rotating}

\definecolor{blue1}{RGB}{0,0, 122}
\usepackage[ruled, linesnumbered]{algorithm2e}
\newcommand{\algcomment}[1]{\textcolor{gray}{\texttt{\# #1}}}

\usepackage{wrapfig,lipsum,booktabs}

\usepackage[accsupp]{axessibility}  

\begin{document}
\pagestyle{headings}
\mainmatter

\title{Unified Fully and Timestamp Supervised Temporal Action Segmentation via Sequence to Sequence Translation}

\titlerunning{Action Segmentation via Seq2Seq Translation}
%
\author{Nadine Behrmann\inst{1, *}
\and
S. Alireza Golestaneh\inst{1, *}
\and
Zico Kolter\inst{1}
\and \\
Juergen Gall\inst{2}
\and
Mehdi Noroozi\inst{1}
}
\authorrunning{N. Behrmann et al.}
%
\institute{Bosch Center for Artificial Intelligence \and
University of Bonn, Germany
}

\newcommand{\mehdi}[1]{\textcolor{red}{Mehdi: #1}}
\newcommand{\nadine}[1]{\textcolor{orange}{#1}}
\newcommand{\alireza}[1]{\textcolor{gray}{Alireza: #1}}
\newcommand{\todo}[1]{\textcolor{red}{TODO: #1}}
\newcommand{\updated}[1]{\textcolor{blue}{#1}}

\newcommand{\eg}{\textit{e.g.,~}}
\newcommand{\ie}{\textit{i.e.,~}}
\renewcommand{\paragraph}[1]{\noindent \textbf{#1}\,}
\renewcommand{\emph}[1]{\textit{#1}}
\maketitle

\begin{abstract}
This paper introduces a unified framework for video action segmentation via sequence to sequence (seq2seq) translation in a fully and timestamp supervised setup. In contrast to current state-of-the-art frame-level prediction methods, we view action segmentation as a seq2seq translation task, \textit{i.e.,} mapping a sequence of video frames to a sequence of action segments. 
Our proposed method involves a series of modifications and auxiliary loss functions on the standard Transformer seq2seq translation model to cope with long input sequences opposed to short output sequences and relatively few videos. We incorporate an auxiliary supervision signal for the encoder via a frame-wise loss and propose a separate alignment decoder for an implicit duration prediction. Finally, we extend our framework to the timestamp supervised setting via our proposed constrained k-medoids algorithm to generate pseudo-segmentations.
Our proposed framework performs consistently on both fully and timestamp supervised settings, outperforming or competing state-of-the-art on several datasets. 
Our code is publicly available at \href{https://github.com/boschresearch/UVAST}{\texttt{https://github.com/boschresearch/UVAST}}.
\let\thefootnote\relax\footnotetext{*  Equal contribution.}
\end{abstract}

\section{Introduction}
\label{sec:intro}

\begin{figure}[t]
\begin{center}
\begin{minipage}[t]{\textwidth}
\includegraphics[width=\linewidth]{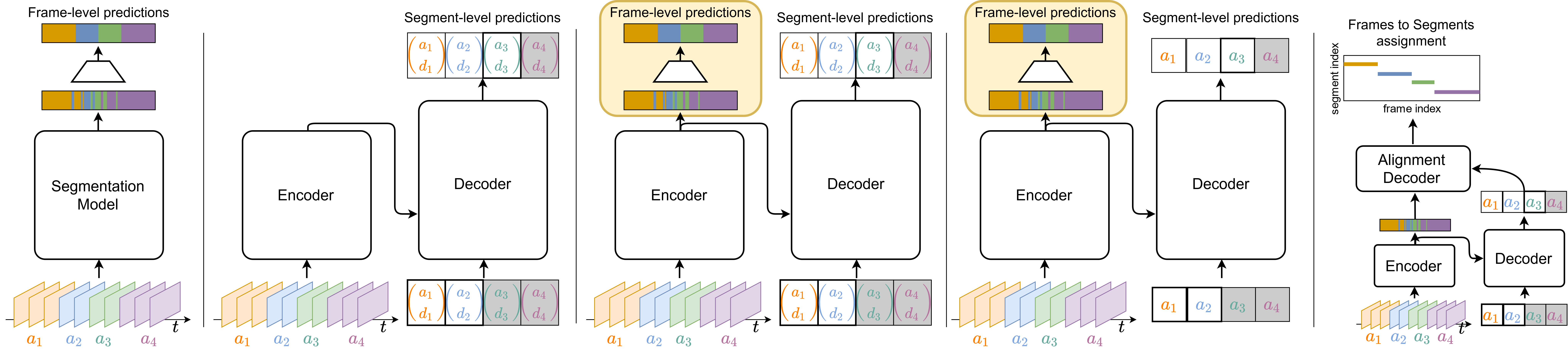}
\end{minipage}

\begin{minipage}[t]{0.12\textwidth}
\centering
(a)
\end{minipage}
\begin{minipage}[t]{0.24\textwidth}
\centering
(b)
\end{minipage}
\begin{minipage}[t]{0.24\textwidth}
\centering
(c)
\end{minipage}
\begin{minipage}[t]{0.24\textwidth}
\centering
(d)
\end{minipage}
\begin{minipage}[t]{0.12\textwidth}
\centering
(e)
\end{minipage}

\end{center}
\caption{
\textbf{Using Transformers for Action Segmentation.}
Instead of frame-level predictions, which are prone to over-segmentation (a), we propose a seq2seq transformer model for segment-level predictions (b). To provide more direct feedback to the encoder we apply a frame-wise loss (c); the resulting features enhance the decoder predictions. However, duration prediction still suffers, so we focus on transcript prediction (d) and use a separate alignment decoder to fuse encoder and decoder features to arrive at an implicit form of duration prediction (e). 
}
\label{fig:model_step_by_step}
\end{figure}

The ability to analyze, comprehend, and segment video content at a temporal level is crucial for many computer vision, video understanding, robotics, and surveillance applications.
Recent state-of-the-art methods for action segmentation mainly formalize the task as a frame-wise classification problem; that is, the objective is to assign an action label to each frame, based upon the full sequence of video frames. We illustrate this general approach in Fig.~\ref{fig:model_step_by_step}~(a). However, this formulation suffers several drawbacks, such as over-segmentation when trained on relatively small datasets (which typically need to consist of expensive frame-level annotations). 

In this work, we propose an alternative approach to the action segmentation task. Our approach involves a transformer-based seq2seq architecture that aims to map from the video frames directly to a \emph{higher-level sequence} of action segments, \textit{i.e.,} a sequence of action label / duration pairs that describes the full predicted segmentation.

The basic structure of our model follows traditional Transformer-based seq2seq models: the encoder branch takes as input a sequence of video frames and maps them to a set of features with the same length; the decoder branch then takes these features as input and generates a predicted sequence of high-level action segments in an auto-regressive manner. 
This approach, illustrated in Fig.~\ref{fig:model_step_by_step}~(b), is a natural fit for action segmentation because it allows the decoder to directly output sequences in the higher-level description space. The main advantage over the frame-level prediction is that it is less prone to over-segmentation.

However, this seemingly natural approach does not immediately perform well on the action segmentation task by itself. In contrast to language translation, action segmentation typically involves long input sequences of very similar frames opposed to short output sequences of action segments. This difference together with the relatively small amount of training videos, makes it challenging for the encoder and decoder to keep track of the full information flow that is necessary to predict the high-level segmentation alone.  
For this reason, we incorporate several modifications and additional loss terms into our system, which together make this approach compete with or improve upon the state-of-the-art.

First, to provide more immediate feedback to the encoder, we employ a frame-wise loss that linearly classifies each frame with the corresponding action label given the encoder features, Fig.~\ref{fig:model_step_by_step}~(c). As a result, the encoder performs frame-wise classification with high localization performance, \textit{i.e.,} high frame-wise accuracy, but low discrimination performance, \textit{i.e.,} over-segmentation with low Edit distance to the ground truth. Nonetheless, its features provide the decoder an informative signal to predict the sequence of actions more accurately. This immediate auxiliary supervision signal allows the decoder to learn more discriminative features for different actions. While the frame-wise loss improves the transcript prediction, the decoder still suffers from low localization performance for duration prediction. As the next step, we fuse the decoder predictions with the encoder, for which we propose two solutions. First, we propose to fuse the discriminative features of the decoder with the encoder features via a cross-attention mechanism in an alignment decoder, Fig.~\ref{fig:model_step_by_step}~(d,e). Second, the high performance of our decoder on predicting transcripts and the high performance of our encoder on localizing actions allows us to effectively utilize the common post-processing algorithm such as FIFA~\cite{fifa2021} and Viterbi~\cite{richard2018nnviterbi,8585084}. 

Finally, we further extend our proposed framework when only a weaker form of timestamp supervision is available. As mentioned before, the frame-wise prediction is vital in our Transformer model to cope with small datasets and long sequences of frames. In this case, when the frame-level annotations are not fully available, we assign a label to each frame by a constrained k-medoids clustering algorithm that takes advantage of timestamp supervision.  Our simple proposed clustering method achieves a frame-wise accuracy of up to $81\%$ on the training set, which can be effectively used to train our seq2seq model. 
We further show that the clustering method can also be used in combination with frame-wise prediction methods such as ASFormer~\cite{asformer}.

We evaluate our model on three challenging action segmentation benchmarks: 50Salads~\cite{stein2013combining}, GTEA~\cite{fathi2011learning}, and Breakfast~\cite{kuehne2014language}. While our method achieves competitive frame-wise accuracies compared to the state-of-the-art, our method substantially outperforms other approaches in predicting the action sequence of a video, which is measured by the Edit distance. By using Viterbi~\cite{richard2018nnviterbi,8585084} or FIFA~\cite{fifa2021} as post-processing, our approach also achieves state-of-the-art results in terms of segmental F1 scores.
To the best of our knowledge, this work is the first that utilizes Transformers in an auto-regressive manner for action segmentation and is applicable to both the fully and timestamp supervised setup.

\section{Related Work}

\textbf{Fully Supervised Action Segmentation.}
Early approaches for action segmentation are based on sliding window and non-maximum suppression \cite{Rohrbach:2012,Karaman:2014}. Other traditional approaches use hidden Markov Models (HMM) for high-level temporal modeling~\cite{kuehne2016end,Tang:2012:latent}. 
\cite{richard2016temporal} use a language and length model to model the probability of action sequences and convert the frame-wise probabilities into action segments using dynamic programming.

More recent approaches are based on temporal convolutions:
\cite{lea2017temporal} propose temporal convolutional networks (TCN) with temporal pooling to capture long-range dependencies within the video. However, such temporal pooling operations struggle to maintain fine-grained temporal information. Therefore, \cite{farha2019ms,li2020ms} use multi-stage TCNs, which maintain a high temporal resolution, with a smoothing loss and refinement modules.
These methods solve the action segmentation task by predicting an action class for each frame, which is prone to over-segmentation and requires refinement modules and smoothing or expensive inference algorithms.
\cite{asrf} address this issue by adding a boundary regression branch to detect action boundaries, which are used during inference to refine the segmentation.
\cite{huang2020improving} propose a graph-based temporal reasoning module that can be built on top of existing methods to refine predicted segmentations.
In contrast, we aim to predict the high-level sequence of segments directly. 

\paragraph{Weakly Supervised Action Segmentation.}
To avoid the costly frame-wise annotations, many methods have been proposed that rely on a weaker form of supervision \cite{bojanowski:weakly:2014,richard2018action,Li:2019:CDFL,Souri:2021:MuCon,ding2018weakly}, such as transcript supervision~\cite{bojanowski:weakly:2014}: Here, only the ordered sequence of actions in the video are given.
\cite{Huang:2016:ECTC} extend the connectionist temporal classification framework, originally introduced for speech recognition, to videos to efficiently evaluate all possible frame-to-action alignments. 
\cite{Ding:2018:ISBA} propose an iterative soft boundary assignment strategy to generate frame-wise pseudo-labels from transcripts. 
\cite{richard2018nnviterbi} generate frame-wise pseudo-labels with the Viterbi algorithm.
\cite{Li:2019:CDFL} extend this work by adding a loss that discriminates between valid and invalid segmentations. 
\cite{Souri:2021:MuCon} use a two-branch neural network with a frame classification branch and a segment generation branch and enforce the two representations to be consistent via a mutual consistency loss. Similar to our method, their segment generation branch also predicts the transcript in an auto-regressive manner and achieves high Edit scores, validating our aspiration for segment-level predictions.
While transcript supervision reduces the annotation cost significantly, the performance suffers. As an alternative, timestamp supervision~\cite{timestamp2021} has been proposed, where for each action segment a single frame is annotated. The annotation cost for such timestamps is comparable to transcript annotations~\cite{timestamp2021} but provides stronger supervision as it gives information about the rough location of the segments.

\paragraph{Transformers.}
Transformers~\cite{Vaswani:attention:2017} originally emerged in the field of natural language processing, and solely rely on the attention mechanism to capture contextual information from the entire sequence. Recently, Transformers have also seen wide adoption in vision-related tasks, \eg image classification~\cite{vit2021}, segmentation~\cite{Zheng:2021:rethinking,Wang:2021:panoptic} and action classification~\cite{Bertasius:2021:TimeSformer,Arnab:2021:vivit}. 
Current standard Transformer-based models are unable to process very long sequences~\cite{beltagy2020longformer,tay2021long,nawrot2021hierarchical,dai2019transformer,zhu2021long}.
One reason for this is the self-attention operation, which scales quadratically with the sequence length.
\cite{beltagy2020longformer} showed that using sliding window attention can reduce the time and memory complexity of the Transformer while preserving the performance. 
Recently, ASFormer~\cite{asformer} leveraged multi-stage TCNs \cite{farha2019ms} and transformer-based models for action segmentation. For each dilated temporal convolutional layer of MS-TCN, an additional self-attention block with instance normalization is added. The first stage is the encoder while the later stages are the decoders, which take the concatenated features of the encoder and the features at the end of the previous stage as input. While we use a similar encoder as ASFormer~\cite{asformer}, our decoder is very different. While ASFormer and MS-TCN perform frame-level prediction as illustrated in Fig.~\ref{fig:model_step_by_step}~(a), our decoder predicts the action segments in an auto-regressive manner as illustrated in Fig.~\ref{fig:model_step_by_step}~(d,e).

\section{Method}
In this section, we introduce our \textbf{U}nified \textbf{V}ideo \textbf{A}ction \textbf{S}egmentation model via \textbf{T}ransformers (\textit{\textbf{UVAST}}).  
The goal of action segmentation is to temporally segment long, untrimmed videos and classify each of the obtained segments. Current state-of-the-art methods are based on \textit{frame-level} predictions -- they assign an action label to each individual frame -- which are prone to \textit{over-segmentation}: The video is not accurately segmented into clean, continuous segments, but fragmented into many shorter pieces of alternating action classes.
We challenge this view of frame-level predictions and propose a novel approach that directly predicts the segments. 
By focusing on \textit{segment-level} predictions -- an alternative but equivalent representation of segmentations -- our method overcomes the deep-rooted over-segmentation problem of frame-level predictions.

\subsection{Transformer for Auto-Regressive Segment Prediction}
\label{sec:model1} 
In this work, we view action segmentation from a sequence-to-sequence (seq2seq) perspective: mapping a sequence of video frames to a sequence of action segments, \textit{e.g.,} as pairs of action label and segment duration. 
The Transformer model~\cite{Vaswani:attention:2017} has emerged as a particularly powerful tool for seq2seq tasks and may seem like the natural fit.
The vanilla Transformer model consists of an encoder module that captures long-range dependencies within the input sequence and a decoder module that translates the input sequence to the desired output sequence in an auto-regressive manner. In contrast to language translation tasks, action segmentation faces a strong mismatch between input and output sequence lengths, \textit{i.e.,} inputs are long and untrimmed videos with various sequence lengths, while outputs are relatively short sequences of action segments. Therefore, we incorporate several modifications to address these issues, which we will go over in more detail in the following.
\newline
\paragraph{Notation.} 
Given an input sequence of $T$ frame-wise features $x_t$, for frame $t\in\{1,\dots,T\}$, our goal is to temporally segment and classify the $T$ frames. The ground-truth labels of a segmentation can be represented in two equivalent forms:
1) a sequence of frame-wise action labels $\hat{y_t}\in\mathcal{C}$ for frame $t$, where $\mathcal{C}$ is the set of action classes,
2) a sequence of segment-wise annotations, which consists of ground-truth segment action classes  $\hat{a}_i\in\mathcal{C}$ (also known as \textit{transcript}), and segment durations $\hat{u}_i\in\mathbb{R}_+$ for each segment $i\in\{1,\dots,N\}$. \\
\paragraph{Transformer Encoder.}
Our input sequence $X\in\mathbb{R}^{T\times d}$ consists of $T$ frame-wise features $x_t$, where $d$ denotes the feature dimension. We embed them using a linear layer and then feed them to the Transformer encoder, which consists of several layers and allows the model to capture long-range dependencies within the video via the self-attention mechanism. The output of the encoder, $E\in\mathbb{R}^{T\times d'}$, is a sequence of frame-wise features $e_t$, which will be used in the cross-attention module of the decoder. To provide direct feedback to the encoder, we apply a linear layer to obtain frame-level predictions for $e_t$. This enables the encoder to accurately localize the action classes within the video and provides more informative features to the decoder.
In practice, we use a modified version of the encoder proposed in \cite{asformer}, which locally restricts the self-attention mechanism and uses dilated convolutions (see supplemental material for more details).\\
\paragraph{Transformer Decoder.}
Given a sequence of frame-wise features $E\in\mathbb{R}^{T\times d'}$, we use a Transformer decoder to auto-regressively predict the transcript, \ie the action labels of the segments.  
Starting with a \textit{start-of-sequence (sos)} token, we feed the sequence of segments $S\in\mathbb{R}^{N\times d'}$ -- embedded using learnable class tokens and positional encoding -- up until segment $i$ to the decoder.  
Via the cross-attention between the current sequence of segments and frame-wise features, the decoder determines the next segment $i+1$ in the video. 
In principle, the decoder could predict the segment duration as well (Fig.~\ref{fig:model_step_by_step}~(c)), however, in practice we found that the decoder's duration prediction suffers from low localization performance, see Table~\ref{tab:expl_dur}. While it is sufficient to pick out a single or few frames in the cross-attention mechanism for predicting the correct action class of a segment, the duration prediction is more difficult since it requires to assign frames to a segment and count them. Since the number of segments is much smaller than the number of frames, the cross-attention mechanism tends to assign only a subset of the frames to the correct segment.    
To address this issue, we propose a separate decoder module, which fuses the discriminative decoder features with the highly localized encoder features to obtain a more accurate duration prediction, which we describe in Section~\ref{sec:alignment_decoder}.

\subsection{Training Objective}
\label{sec:losses}

Although our ultimate goal is segment-level predictions, we provide feedback to both the encoder and decoder model to make the best use of the labels.
To that end, we apply a frame-wise cross-entropy loss on the frame-level predictions of the encoder:
\begin{align}
    \mathcal{L}_{\text{frame}}&=-\frac{1}{T}\sum_{t=1}^{T}\log(y_{t,\hat{c}}),
\label{eq:ce_frame}
\end{align}
where $y_{t,c}$ denotes the predicted probability of label $c$ at time $t$, and $\hat c$ denotes the ground-truth label of frame $t$.
Analogously, we apply a segment-wise cross-entropy loss on the segment-level predictions of the decoder:
\begin{align}
    \mathcal{L}_{\text{segment}}&=-\frac{1}{N}\sum_{i=1}^{N}\log(a_{i,\hat{c}}),
\label{eq:ce_segment}
\end{align}
where $a_{i,c}$ denotes the predicted probability of label $c$ at segment $i$, and $\hat c$ denotes the ground-truth label of segment $i$.
\newline
\paragraph{Regularization via Grouping.} To regularize the encoder and decoder predictions, we additionally apply \textit{group-wise} cross-entropy losses. To that end, we group the frames and segments by ground-truth labels $L = \{c \in \mathcal{C}|c\in \{\hat a_1,\dots,\hat a_n\}\}$ that occur in the video: $T_c = \{t\in \{1,\dots,T\}| \hat y_{t} = c\}$ are the indices of frames with class $c$, and $N_c = \{i\in \{1,\dots,N\}| \hat a_{i} = c\}$ the indices of segments with class $c$.
We apply a cross-entropy loss to the averaged prediction of each group:
\begin{align}
    \mathcal{L}_{\text{g-frame}} &= - \frac{1}{|L|}\sum_{c \in L} \log\left(\frac{1}{|T_c|}\sum_{t\in T_c} y_{t, c}\right)
    \label{eq:g_frame}\\
    \mathcal{L}_{\text{g-segment}} &= - \frac{1}{|L|}\sum_{c \in L} \log\left(\frac{1}{|N_c|}\sum_{i\in N_c} a_{i, c} \right) 
    \label{eq:g_segment}
\end{align}

\subsection{Cross-Attention Loss}
We utilize a loss through a cross-attention mechanism between the encoder and decoder features to allow further interactions between them. Let us assume that $T$ video frames and corresponding $N$ actions in the encoder and decoder are represented by their features $E\in\mathbb{R}^{T\times d^\prime}$ and $D\in\mathbb{R}^{N\times d^{\prime}}$, respectively. The cross-attention loss involves obtaining a cross-attention matrix $M=\texttt{softmax}(\frac{ED^{T}}{{\tau'\sqrt{d'}}})$, where $\tau'$ is a stability temperature, and each row of $M$ includes a probability vector that assigns each encoder feature (frame) to decoder features  (actions). We then use $M$ in the following cross-entropy loss function:
\begin{align}
\label{eq:ca}
    \mathcal{L}_{\text{CA}}(M) = -\frac{1}{T} \sum_t \log(M_{t,\hat n}), 
\end{align}
where $\hat n$ is the ground-truth segment index to which frame $t$ belongs. 
We use this loss in our transcript decoder (main decoder) and alignment decoder in the following.

\paragraph{Cross-Attention Loss for the Transcript Decoder.}
The cross-attention loss, when applied to the transcript decoder, provides more intermediate feedback to the decoder about the action location in the input sequence, see Fig.~\ref{fig:ca_ablation}. We found this loss function especially effective on smaller datasets such as 50Salads (see Table~\ref{tab:ablation_losses}). 
Our main objective for the encoder and the transcript decoder is:
\begin{align}
    \mathcal{L} = \mathcal{L}_{\text{frame}} + \mathcal{L}_{\text{segment}} + \mathcal{L}_{\text{g-frame}} + \mathcal{L}_{\text{g-segment}} + \mathcal{L}_{\text{CA}}(M),
    \label{eq6}
\end{align}

\begin{figure}[t]
    \centering
    \includegraphics[width=\linewidth]{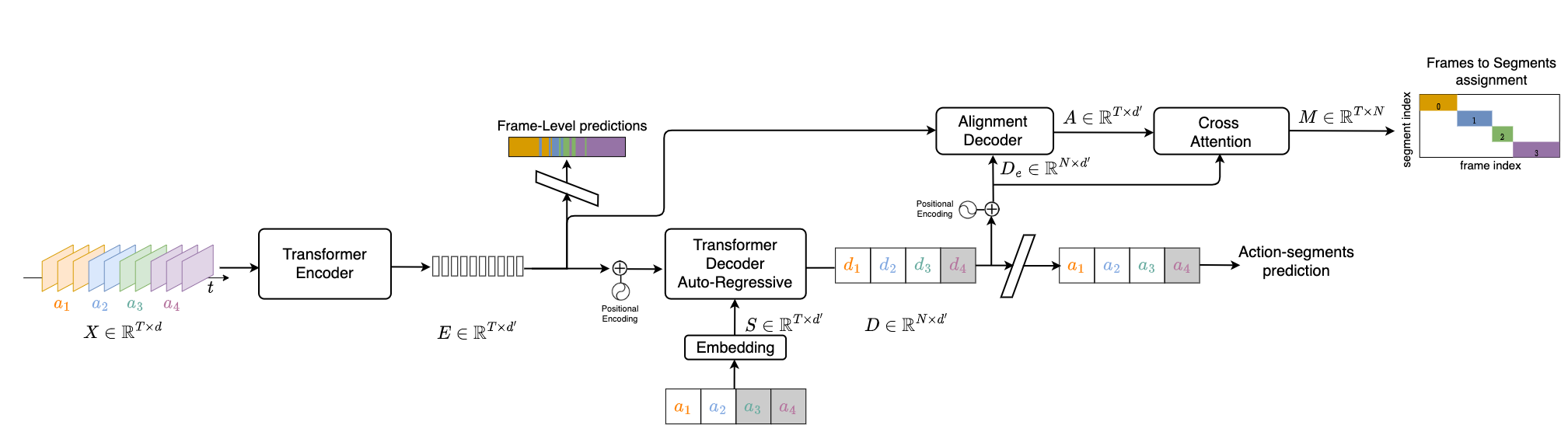}
    \caption{
    \textbf{Overview of our complete model.}
    Our complete model consists of a Transformer encoder and an auto-regressive Transformer decoder, which we train for frame-level and segment-level predictions, respectively. For duration prediction we use an alignment decoder -- followed by cross attention -- on top of the encoder and decoder features to compute a frames-to-segment assignment, which is used to compute the durations of the segments.
    }
    \label{fig:full_model}
\end{figure}

\paragraph{Cross-Attention Loss for the Alignment Decoder.}
\label{sec:alignment_decoder}
While the transcript decoder generates the sequence of actions in a video, it does not predict the duration of each action. Although it is possible to predict the duration as well, as illustrated in Fig.~\ref{fig:model_step_by_step}~(c), the transcript decoder still struggles to localize actions through direct duration prediction as shown in Table~\ref{tab:expl_dur}. One reason for this could be the high mismatch between input and output sequence length and the relatively small number of training videos. While picking up a single segment frame is sufficient to predict the action class, the duration prediction effectively requires counting the number of frames in the segment, resulting in a more challenging task. Therefore, we design an alternative alignment decoder for predicting segment durations implicitly. The full flow of the complete model is shown in Fig.~\ref{fig:full_model}.

A high Edit score of our decoder indicates that it has already learned discriminative features of the actions. The motivation for our alignment decoder is to align the encoder features to the highly discriminative features of the decoder, which can be further used for the duration prediction (see Fig~\ref{fig:model_step_by_step}~(e)). In essence, our proposed alignment decoder is a one-to-many mapping from the decoder features to the encoder features. 
The alignment decoder takes the encoder and decoder features $E\in\mathbb{R}^{T\times d^\prime}$ and $D\in\mathbb{R}^{N\times d^{\prime}}$ with positional encoding as input and generates the aligned features $A\in\mathbb{R}^{T\times d'}$. Since the alignment decoder aims to explore the dependencies between the encoder features and the decoder features, we employ a cross-attention mechanism in its architecture similar to the transcript decoder.   
To this end, we compute an assignment matrix $\overline{M}\in\mathbb{R}^{T\times N}$ via cross-attention between the alignment decoder features ($A$) and positional encoded features of the transcript decoder ($D$) by $\overline{M}=\texttt{softmax}(\frac{A{D}^{T}}{\tau})$ with a small value of $\tau$. Note that with a small value of $\tau$ each row of $\overline{M}$ will be close to a one-hot-encoding indicating the segment index the frame is assigned to. The positional encoding for $D$ resolves ambiguities if the same action occurs at several locations in the video. 

In contrast to the decoder from the previous section, the alignment decoder is not auto-regressive since the full sequences of frame-wise and segment-wise features are already available from the previous encoder and decoder. During inference, we compute the segment durations by taking the sum over the assignments:
\begin{align}
\label{eq:ca2}
    u_i = \sum_t \overline{M}_{t, i}, 
\end{align}
where $i\in \{1,...,n\}$ and $\overline{M}_{t, i}$ denotes whether frame $t$ is assigned to segment $i$. We found that training the alignment decoder using only the loss for $\overline{M}$ \eqref{eq:ca2} in a separate stage on top of the frozen encoder and decoder features results in a more robust model that suffers less from overfitting.

\begin{figure}[t]
\begin{minipage}{.63\linewidth}
\begin{algorithm}[H]\small
{\textbf{Input:} $T$ features $x_t$, timestamps $[t_1,\dots,t_n]$}\\
{\textbf{Init:} $m_i = x_{t_i}$ \algcomment{initialize medoids}}\\
\Repeat{until convergence}{
    $D_{i, j}=\text{dist}(m_i, x_j)$ \algcomment{pairwise costs} \\
    $b_0=0; b_n = T$ \algcomment{compute boundaries}\\
    \For{$i=1,\dots,n-1$}{
    $b_i = \text{argmin}_{l}(\sum_{j=t_i}^lD_{i,j} + \sum_{j=l+1}^{t_{i+1}}D_{i+1,j})$\\
    }
    \For{$i=1,\dots,n$}{
    $t_i = \text{argmin}_l(\sum_{j=b_{i-1}+1}^{b_{i}} \text{dist}(x_l, x_j))$ \\
    $m_i = x_{t_i}$ \algcomment{new medoids}
    }
}
\Return $l_i = b_i - b_{i - 1}$
\caption{\footnotesize{Constrained K-medoids to generate temporally continuous clusters.}} 
\label{alg:kmethoids}
\end{algorithm}
\end{minipage}
\begin{minipage}{.36\linewidth}
    \centering
    \includegraphics[width=\linewidth]{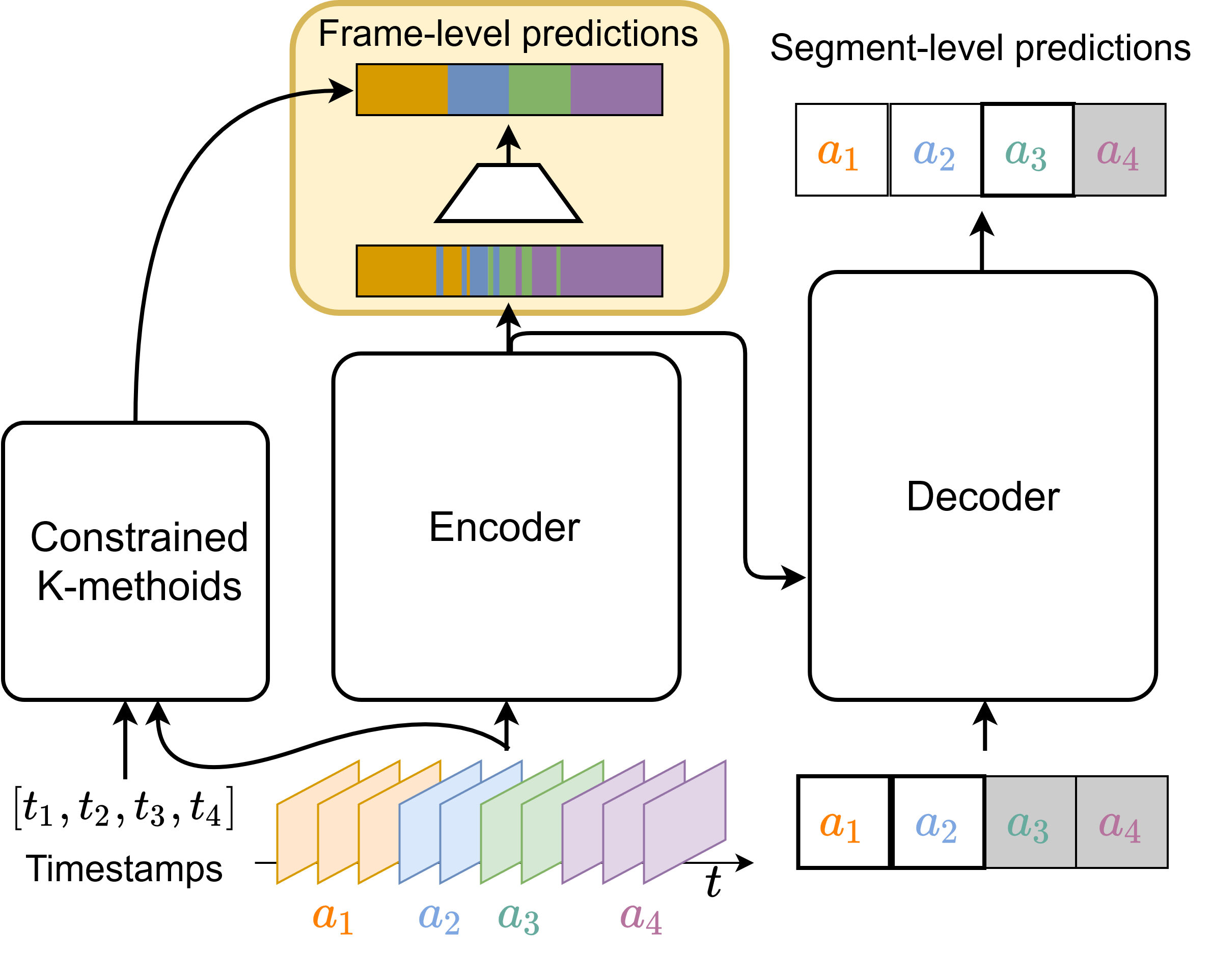}
    \caption{\textbf{Constrained K-medoids.} Given frame-wise features and timestamps, k-medoids generates a pseudo-segmentation that guides the encoder during the training instead of ground truth frame-level labels in a fully supervised setup.}
    \label{fig:kmethoids_timestamps}
\end{minipage}
\end{figure}

\subsection{Timestamp Supervision}
\label{sec:kmethoids_timestamp}
In this section, we show how our proposed framework can be extended to the timestamp supervised setting.
In this setting, we are given a single annotated frame for each segment in the video, \textit{i.e.,} frame annotations are reduced dramatically, and ground-truth segment durations are no longer available for all frames. As we extensively discussed before, our proposed framework relies on the frame-level supervisory signal on top of the encoder. However, it turns out that a noisy frame-level annotation provides a solid signal to the encoder. To obtain such frame-level annotations, we propose a constrained k-medoids algorithm that propagates the timestamp supervision to all frames.

A typical k-medoids algorithm starts with random data points as the cluster centers. It iteratively updates the cluster centers chosen from the data points and the assignments based on their similarity to the cluster center. Having access to the timestamp supervision, we can use them as initialization and cluster the input features. However, in a standard k-medoids algorithm, a temporally continuous set of clusters are not taken for granted. We call our method constrained k-medoids because we force the clusters to be temporally continuous. This can be simply achieved by modifying the assignment step of the k-medoids algorithm. Instead of assigning pseudo-labels to each frame, we find the temporal boundaries of each cluster. In the assignment step, we update the boundaries such that the accumulative distance of each cluster to the current center is minimized. 
Alg.~\ref{alg:kmethoids} summarizes the steps of our clustering method.
In principle, we can apply k-medoids using the frame-wise input features $x_t$, the encoder features $e_t$, or a combination of both. 
In practice, we found that using input features alone gives surprisingly accurate segmentations, see Table~\ref{table:kmethoids} or supplemental material for more analyses.

\section{Experiments}

\subsection{Datasets}
We evaluate the performance of our proposed model extensively on three challenging action segmentation datasets (50Salads \cite{stein2013combining}, GTEA \cite{fathi2011learning}, and Breakfast \cite{kuehne2014language}).  
We follow previous work~\cite{asrf,asformer,li2020ms,farha2019ms,wang2020boundary,chen2020action} and perform 4-fold cross-validation on Breakfast and GTEA and 5-fold cross-validation on 50Salads.

\subsection{Evaluation Metrics}
For evaluation,  following previous works, we report the  frame-wise accuracy (Acc), segmental edit score (Edit), and the segmental F1 score at overlapping thresholds $10\%$, $25\%$, and $50\%$, denoted by $\text{F1@}\{10, 25, 50\}$~\cite{lea2017temporal}. The overlapping threshold is determined based on the intersection over union (IoU) ratio. 
Although frame-wise accuracy is the most commonly used metric for action segmentation, it does not portray a thorough picture of the performance of action segmentation models. 
A major disadvantage of frame-wise accuracy is that long action classes have a higher impact than short action classes and dominate the results. Furthermore, over-segmentation errors have a relatively low impact on Acc, which is particularly problematic for applications such as video summarization.  
On the other hand, Edit and F1 scores establish more comprehensive measures of the quality of the segmentations~\cite{lea2017temporal}; Edit measures the quality of the predicted transcript of the segmentation, while F1 scores penalize over-segmentation and are also insensitive to the duration of the action classes. Our proposed method performs particularly well on the Edit, and F1 scores on all datasets and in fully and timestamp supervised setups, achieving state-of-the-art results in most cases.

\subsection{Implementation Details and Training}
We follow the standard training strategy from existing  algorithms \cite{farha2019ms,li2020ms,asrf,asformer,singhania2021coarse} and train our main network (Section~\ref{sec:model1}) end-to-end with batch size of $1$. We train our
model for at most $800$ epochs using Adam optimizer with learning rate $0.0005$ and the loss (\ref{eq6}). 
In the cross-attention loss, Eq.~\eqref{eq:ca2}, we set $\tau=1$ during training to ensure training stability, and $\tau=0.0001$ during inference.
As input for our model, we use the same I3D \cite{carreira2017quo} features that
were used in many previous works.
For the encoder, we used a modified version of the encoder proposed in~\cite{asformer}\footnote{Please see the supplementary material for more details, hyper-parameters, and ablations.\label{footnote1}}.
For the decoder, we use a standard decoder architecture~\cite{Vaswani:attention:2017}, with two layers and single head attention.
Due to a strong imbalance in the segment durations, we propose a \textit{split-segment} approach for improved training: longer action segments are split up into several shorter ones so that segment durations are more uniformly distributed; for details and ablations, see supplemental material. During the inference, we do not use any split-segment and use the entire video.

For the alignment decoder (Section~\ref{sec:alignment_decoder}), we use a single layer, single head decoder. To train this model, we use similar hyper-parameters and optimizers while freezing the encoder-decoder model from Section~\ref{sec:model1} and only train the alignment decoder with our cross-attention loss. For positional encoding, we use the standard sinusoidal positional encoding~\cite{Vaswani:attention:2017}.
Furthermore, we use random dropping of the features as an augmentation method, where we randomly drop $\sim1\%$ of the features in the sequence.

\subsection{Performance Evaluation}
Here, we provide the overall performance comparison of our proposed method, \textit{\textbf{UVAST}}, on three challenging action segmentation datasets with different levels of supervision.
We demonstrate the effectiveness of our proposed method for both the fully supervised and timestamp supervised setup and achieve competitive results on both settings.
We provide the results of our proposed model for four scenarios: Transcript prediction of our encoder-decoder architecture (referred to as ``w/o duration") and three different approaches to obtain durations for the segments, namely alignment decoder from Section~\ref{sec:alignment_decoder} (``+ alignment decoder"), Viterbi (``+ Viterbi"), and FIFA~\cite{fifa2021} (``+ FIFA").
We only report the Edit score for ``w/o duration", as it does not provide segment durations. 
A significant advantage of our method is that a predicted transcript is readily available and can be used in these inference algorithms instead of the previous methods, which need to iterate over the training transcripts. Furthermore, we can optionally use the predicted duration of the alignment decoder to initialize the segment lengths in FIFA. 

\subsubsection{Fully Supervised Comparison.}
Table~\ref{tab:fully_supervised} shows the performance of our method in the fully supervised setting compared with state-of-the-art methods.
At the bottom of Table~\ref{tab:fully_supervised} we provide the results of our proposed model for the four scenarios explained above. 
\textit{\textbf{UVAST}} achieves significantly better Edit score on transcript prediction (``w/o duration") than all other existing methods on all three datasets, which demonstrates the effectiveness of our model to capture and summarize the actions occurring in the video. 
\begin{table}[t]
\centering
\caption{\textbf{Fully supervised results on all three datasets.} Best and second best results are shown in bold and underlined, respectively. With the assistance of Viterbi/FIFA our method outperforms state-of-the-art in terms of Edit and F1 scores on all datasets.}
\resizebox{\textwidth}{!} {
\begin{tabular}{ll|ccc|c|c||ccc|c|c||ccc|c|c}
\hline 
\hline 
\multicolumn{2}{l|}{} & \multicolumn{5}{c||}{{Breakfast}} & \multicolumn{5}{c||}{{50Salads}} & \multicolumn{5}{c}{{GTEA}}\\
\cline{3-17} 
 &  & \multicolumn{3}{c|}{{F1@\{10,25,50\}}} & Edit & Acc & \multicolumn{3}{c|}{{F1@\{10,25,50\}}} & Edit & Acc & \multicolumn{3}{c|}{{F1@\{10,25,50\}}} & Edit & Acc \\
\hline 
\multicolumn{2}{l|}{TDRN \cite{lei2018temporal}} & - & - & - & - & - & $72.9$ & $68.5$ & $57.2$ & $66.0$ & $68.1$ & $79.2$ & $74.4$ & $62.7$ & $74.1$ & $70.1$ \\
\multicolumn{2}{l|}{SSA-GAN \cite{gammulle2020fine}} & - & - & - & - & - & $74.9$ & $71.7$ & $67.0$ & $69.8$ & $73.3$ & $80.6$ & $79.1$ & $74.2$ & $76.0$ & $74.4$\\
\multicolumn{2}{l|}{MuCon \cite{Souri:2021:MuCon}} & $73.2$ & $66.1$ & $48.4$ & $76.3$ & $62.8$ & - & - & - & - & - & - & - & - & - & -\\
\multicolumn{2}{l|}{{DTGRM \cite{wang2020temporal}}} & $68.7$ & $61.9$ & $46.6$ & $68.9$ & $68.3$ & $79.1$ & $75.9$ & $66.1$ & $72.0$ & $80.0$ & $87.3$ & $85.5$ & $72.3$ & $80.7$ & $77.5$\\
\multicolumn{2}{l|}{Gao  \textit{et al.} \cite{gao2021global2local}} & $74.9$ & $69.0$ & $55.2$ & $73.3$ & $70.7$ & $80.3$ & $78.0$ & $69.8$ & $73.4$ & $82.2$ & $89.9$ & $87.3$ & $75.8$ & $84.6$ & $78.5$\\
\multicolumn{2}{l|}{MS-TCN++ \cite{li2020ms}} & $64.1$ & $58.6$ & $45.9$ & $65.6$ & $67.6$ & $80.7$ & $78.5$ & $70.1$ & $74.3$ & $83.7$ & $88.8$ & $85.7$ & $76.0$ & $83.5$ & $80.1$\\
\multicolumn{2}{l|}{BCN \cite{wang2020boundary}} & $68.7$ & $65.5$ & $55.0$ & $66.2$ & $70.4$ & $82.3$ & $81.3$ & $74.0$ & $74.3$ & $84.4$ & $88.5$ & $87.1$ & $77.3$ & $84.4$ & $79.8$\\
\multicolumn{2}{l|}{SSTDA \cite{chen2020action}} & $75.0$ & $69.1$ & $55.2$ & $73.7$ & $70.2$ & $83.0$ & $81.5$ & $73.8$ & $75.8$ & $83.2$ & $90.0$ & $89.1$ & $78.0$ & $86.2$ & $79.8$\\
\multicolumn{2}{l|}{Singhania \textit{et al.} \cite{singhania2021coarse}} & $70.1$ & $66.6$ & $56.2$ & $68.2$ & \uline{$73.5$} & $76.6$ & $73.0$ & $62.5$ & $69.2$ & $80.1$ & $90.5$ & $88.5$ & $77.1$ & $87.3$ & $\mathbf{80.3}$\\
\multicolumn{2}{l|}{ASRF \cite{asrf}} & $74.3$ & $68.9$ & $56.1$ & $72.4$ & $67.6$ & $84.9$ & $83.5$ & $77.3$ & $79.3$ & $84.5$ & $89.4$ & $87.8$ & \uline{$79.8$} & $83.7$ & $77.3$\\
\multicolumn{2}{l|}{ASFormer \cite{asformer}} & $76.0$ & $70.6$ & $57.4$ & $75.0$ & \uline{$73.5$} & $85.1$ & $83.4$ & $76.0$ & \uline{$79.6$} & \uline{$85.6$} & $90.1$ &  $88.8$  & $79.2$ & $84.6$ & $79.7$\\
\multicolumn{2}{l|}{ASFormer \cite{asformer} + Viterbi} & $76.1$ & $70.5$ & $57.1$ & $74.5$ & $70.2$ & $84.1$ & $82.3$ & $74.9$ & $76.1$ & $84.7$ & \underline{$91.1$} & \underline{$90.0$} & $79.5$ & $86.5$ & $80.0$ \\
\multicolumn{2}{l|}{ASFormer \cite{asformer} + FIFA} & \underline{$76.8$} & \underline{$71.4$} & $\mathbf{58.9}$ & $75.6$ & $\mathbf{73.7}$ & $84.5$ & $83.2$ & $75.4$ & $78.5$ & $85.4$ & $90.4$ & $88.6$ & $78.1$ & $86.2$ & $78.9$ \\ 
\hline 
\multirow{4}{*}{UVAST (Ours)} & w/o duration & - & - & - & $76.9$ & - & - & - & - & $\mathbf{83.9}$ & - & - & - & - & $\mathbf{92.1}$ & - \\
  & + alignment decoder & $76.7$ & $70.0$ & $56.6$ & $\mathbf{77.2}$ & $68.2$ & $86.2$ & $81.2$ & $70.4$ & $\mathbf{83.9}$ & $79.5$ & $77.1$ & $69.7$ & $54.2$ & \uline{$90.5$} & $62.2$ \\
  & + Viterbi & $75.9$ & $70.0$ & $57.2$ & $76.5$ & $66.0$ & $\mathbf{89.1}$ & $\mathbf{87.6}$ & $\mathbf{81.7}$ & $\mathbf{83.9}$ & $\mathbf{87.4}$ & $\mathbf{92.7}$ & $\mathbf{91.3}$ & $\mathbf{81.0}$ & $\mathbf{92.1}$ & \uline{$80.2$} \\
  & + FIFA & $\mathbf{76.9}$ & $\mathbf{71.5}$ & \underline{$58.0$} & \uline{$77.1$} & $69.7$ & \uline{$88.9$} & \uline{$87.0$} & \uline{$78.5$} & $\mathbf{83.9}$ & $84.5$ & $82.9$ & $79.4$ & $64.7$ & \uline{$90.5$} & $69.8$ \\
\hline 
\hline
\end{tabular}
}
\label{tab:fully_supervised}
\end{table}
In the last three rows of Table~\ref{tab:fully_supervised}, we use three different approaches to compute the duration of the segments.
Combining \textit{\textbf{UVAST}} with the alignment decoder from Section~\ref{sec:alignment_decoder} achieves competitive results. However, it is important to note that Transformers are very data-hungry and training them on small datasets can be challenging.
We observe that \textit{\textbf{UVAST}} with alignment decoder outperforms other methods in terms of Edit score. While the F1 scores are comparable to the state-of-the-art on the Breakfast dataset, the small size of the GTEA dataset hinders the training of the alignment decoder. 

Moreover, with frame-wise predictions and transcript prediction available, our method conveniently allows applying inference algorithms at test time, such as FIFA and Viterbi, without the need to expensively iterate over the training transcripts.
Combining our method with Viterbi outperforms the existing methods on GTEA and 50Salads in terms of Edit and F1 scores, and achieves competitive results on Breakfast. We also provide the results of \textit{\textbf{UVAST}} with FIFA, where we initialize the duration with the predicted duration. It achieves strong performance on Breakfast and 50Salads. Note that although FIFA is a fast approximation of Viterbi, it achieves better results on the Breakfast dataset. This is due to the fact that the objective function that is minimized by FIFA/Viterbi does not optimize the evaluation metrics directly, \textit{i.e.,} the global optimum of the Viterbi objective function does not guarantee the global optimum of the evaluation metrics. This observation is consistent with the results reported in~\cite{fifa2021}.     

The comparison to ASFormer~\cite{asformer} is also interesting. While ASFormer performs like most other approaches frame-level prediction, Fig.~\ref{fig:model_step_by_step}~(a), \textit{\textbf{UVAST}} predicts the action segments in an autoregressive manner, Fig.~\ref{fig:model_step_by_step}~(d,e). As expected, ASFormer achieves in general a better frame-wise accuracy while \textit{\textbf{UVAST}} achieves a better Edit score. Since ASFormer uses a smoothing loss and multiple refinement stages to address over-segmentation similar to MS-TCN~\cite{farha2019ms,li2020ms}, it has $\sim1.3$M learnable parameters, whereas our proposed model has $\sim1.1$M parameters. Our approach with Viterbi achieves similar F1 scores on the Breakfast dataset, but higher F1 scores on the other datasets.    
For a more thorough and fair comparison with ASFormer, we additionally provide the results when combined with Viterbi or FIFA during inference. To that end, we extract the transcript to be used in Viterbi/FIFA from the frame-wise predictions of the model.

Overall, we find that our method achieves strong performance in terms of Edit and F1 scores, while Acc is compared to the state-of-the-art lower on Breakfast. Note that Acc is dominated by long segments and less sensitive to over-segmentation errors. Lower Acc and higher Edit/F1 scores indicate that \textit{\textbf{UVAST}} localizes action boundaries, which are difficult to annotate precisely, less accurately. It is therefore an interesting research direction to improve the segment boundaries, \textit{e.g.,} by using an additional refinement like ASFormer.      

\begin{table}[t!]
\begin{center}
\caption{
\textbf{Timestamp supervision results on all three datasets.}
UVAST, ASFormer~\cite{asformer}, and MSTCN~\cite{farha2019ms} are trained via our constrained k-medoids pseudo-labels. Best result shown in bold. \textbf{\textit{UVAST}} outperforms SOTA on all datasets and metrics except for Acc on Breakfast. The performance in terms of Edit distance is significant, and is comparable to the fully supervised setup.
}
\label{tab:timestamp}
\resizebox{\linewidth}{!} {
\begin{tabular}{ll|ccc|c|c||ccc|c|c||ccc|c|c}
\hline 
\hline 
\multicolumn{2}{l|}{} & \multicolumn{5}{c||}{{Breakfast}} & \multicolumn{5}{c||}{{50Salads}} & \multicolumn{5}{c}{{GTEA}}\\
\cline{3-17} 
 &  & \multicolumn{3}{c|}{{F1@\{10,25,50\}}} & Edit & Acc & \multicolumn{3}{c|}{{F1@\{10,25,50\}}} & Edit & Acc & \multicolumn{3}{c|}{{F1@\{10,25,50\}}} & Edit & Acc \\
\hline 
\multicolumn{2}{l|}{Li et al. \cite{timestamp2021}} & $70.5$ & $63.6$ & $47.4$ & $69.9$ & $\mathbf{64.1}$ & $73.9$ & $70.9$ & $60.1$ & $66.8$ & $75.6$ & $78.9$ & $73.0$ & $55.4$ & $72.3$ & $66.4$\\
\multicolumn{2}{l|}{MS-TCN~\cite{farha2019ms}} & $56.1$ & $50.0$ & $36.8$ & $61.7$ & $62.5$ & $74.4$ & $70.4$ & $57.7$ & $66.8$ & $72.8$ & $82.8$ & $80.3$ & $63.5$ & $79.5$ & $67.7$\\
\multicolumn{2}{l|}{MS-TCN~\cite{farha2019ms} + Viterbi} & $43.3$ & $37.2$ & $25.6$ & $43.5$ & $35.9$ & $74.0$ & $70.0$ & $55.5$ & $68.2$ & $72.8$ & $82.6$ & $79.7$ & $61.6$ & $81.0$ & $68.1$\\
\multicolumn{2}{l|}{MS-TCN~\cite{farha2019ms} + FIFA} & $36.1$ & $30.5$ & $21.9$ & $42.6$ & $35.8$ & $73.7$ & $69.5$ & $54.9$ & $69.2$ & $72.8$ & $81.7$ & $77.7$ & $57.7$ & $81.0$ & $67.3$ \\
\multicolumn{2}{l|}{ASFormer~\cite{asformer}} & $70.9$ & $62.9$ & $44.0$ & $71.6$ & $61.3$ & $76.6$ & $72.1$ & $59.6$ & $70.0$ & $76.9$ & $\mathbf{87.2}$ & $83.1$ & ${67.5}$ & $83.0$ & $68.8$ \\
\multicolumn{2}{l|}{ASFormer~\cite{asformer} + Viterbi} & $71.3$ & $63.1$ & $44.3$ & $71.1$ & $60.7$ & $76.3$ & $72.1$ & $59.4$ & $68.8$ & $\mathbf{77.0}$ & $87.1$ & $83.1$ & $\mathbf{68.2}$ & $83.0$ & $69.1$ \\
\multicolumn{2}{l|}{ASFormer~\cite{asformer} + FIFA} & $71.1$ & $62.7$ & $44.3$ & $71.8$ & $61.8$ & $76.7$ & $72.0$ & $58.8$ & $70.0$ & $76.9$ & $86.8$ & $81.9$ & $65.4$ & $83.0$ & $68.4$ \\
\hline 
\multirow{3}{*}{{UVAST (Ours)}}
 & + alignment decoder & $\mathbf{72.0}$ & $64.1$ & $\mathbf{48.6}$ & $\mathbf{74.3}$ & $60.2$ & $75.7$ & $70.6$ & $58.2$ & $78.4$ & $67.8$ & $70.8$ & $63.5$ & $49.2$ & $88.2$ & $55.3$\\
 & + Viterbi & $71.3$ & $63.3$ & $48.3$ & $74.1$ & $60.7$ & $\mathbf{83.0}$ & $\mathbf{79.6}$ & $\mathbf{65.9}$ & $78.2$ & $\mathbf{77.0}$ & $\mathbf{87.2}$ & $\mathbf{83.7}$ & $66.0$ & $\mathbf{89.3}$ & $\mathbf{70.5}$\\
 & + FIFA & $\mathbf{72.0}$ & $\mathbf{64.2}$ & $47.6$ & $74.1$ & $60.3$ & $80.2$ & $74.9$ & $61.6$ & $\mathbf{78.6}$ & $72.5$ & $80.7$ & $75.2$ & $57.4$ & $88.7$ & $66.0$\\
\hline 
\hline 
\end{tabular}    
}
\end{center}
\end{table}

\subsubsection{Timestamp Supervision Comparison.}
We use our proposed constrained k-medoids to generate pseudo-segmentation using the frame-wise input features and ground truth timestamps. The output consists of continuous segments, which can be identified with the transcript to yield a pseudo-segmentation. While this approach can be applied both to the input features and encoder features in principle, we find that using the input features already gives a surprisingly good performance; we report Acc and F1 scores in Table~\ref{table:kmethoids} averaged over all splits. Note that this is not a temporal segmentation method as it requires timestamp supervision as input. We use the resulting pseudo-segmentation as the auxiliary signal to our encoder during the training where we have access to the timestamp supervision. 

In Table~\ref{tab:timestamp}, we compare our proposed timestamp model with the recently proposed method~\cite{timestamp2021} on the three action segmentation datasets.
To the best of our knowledge, \cite{timestamp2021} is the first work that proposed and applied timestamp supervision for the temporal action segmentation task. 
Although other weakly supervised methods exist, they are based on \textit{transcript} supervision, a weaker form of supervision; 
therefore, we additionally train MS-TCN~\cite{farha2019ms} and ASFormer~\cite{asformer} with our constrained k-medoids. To get more thorough and fair comparisons, we further show their performance when combined with Viterbi decoding or FIFA during inference.  

Table~\ref{tab:timestamp} shows that: I) our  method largely outperforms the other methods by achieving the best performance on $13$ out of $15$ metrics.
Analogously to the fully supervised case, we observe the strong performance of our alignment decoder in terms of Edit and F1 scores on Breakfast; with FIFA and Viterbi, we outperform the method of \cite{timestamp2021} on 50Salads and GTEA. Notably, \textbf{\textit{UVAST}} achieves significantly higher performance in terms of Edit distance, which is comparable to the fully supervised setup. II) ASFormer and MSTCN perform reasonably well in the timestamp supervision setup when trained on the pseudo-labels of our constrained k-medoids algorithm, which demonstrates one more time the effectiveness of our proposed constrained k-medoids algorithm. III) ASFormer and MSTCN do not benefit from the Viterbi algorithm in this case. This is due to the relatively lower Edit distance of these methods. Namely, Viterbi hurts MSTCN on Breakfast as it achieves significantly lower Edit distance compared to ours.

\subsection{Qualitative Evaluation}
\begin{figure}
\begin{minipage}{\linewidth}
    \centering
    \includegraphics[width=0.92\linewidth]{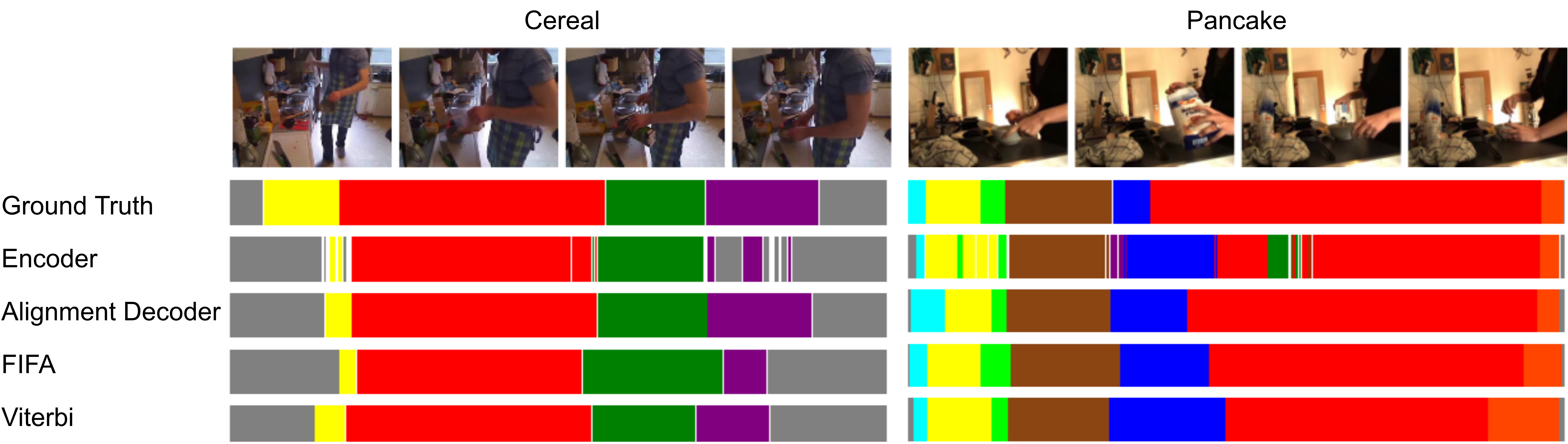}
    \caption{\textbf{Qualitative results.} We show ground truth and predicted segmentation of fully supervised (left) and timestamp supervised (right) \textit{\textbf{UVAST}} of two videos from the Breakfast dataset.}
    \label{fig:qualitative}
\end{minipage}
\end{figure}

We show qualitative results of two videos from the Breakfast dataset in the fully supervised and timestamp supervised setting in  Fig.~\ref{fig:qualitative}. We visualize the ground truth segmentations (first row) as well as the predicted segmentations of our encoder (second row) and decoder with alignment decoder, FIFA or Viterbi for duration prediction (last three rows).
The encoder predictions demonstrate well the common problem of over-segmentation with frame-level predictions; the segment-level predictions of our decoder on the other hand yield coherent action segments.

\subsection{Ablations Studies}
\paragraph{Duration Prediction.}
As discussed in Section~\ref{sec:intro}, the vanilla Transformer model, Fig.~\ref{fig:model_step_by_step}~(b),  does not generalize to the action segmentation task, see Table~\ref{tab:expl_dur}. We train this model using $\mathcal{L}_{\text{segment}}$ and MSE between predicted and ground truth durations, which are scaled to $[0, 1]$ by dividing by the total number of frames $T$.
Our first modification involves applying a frame-wise loss to the encoder features, which drastically improves the results. However, this explicit duration prediction still struggles to accurately localize the segments. 
Predicting duration implicitly via our alignment decoder instead, Fig.~\ref{fig:model_step_by_step}~(d)+(e), on the other hand improves the localization, increasing Acc and F1.

\begin{table}[h]
\centering
\caption{\textbf{Constrained K-medoids results.} We evaluate the pseudo-segmentations of our constrained k-medoids algorithm, Alg.~\ref{alg:kmethoids}, given the frame-wise input features and ground truth timestamps.}
\label{table:kmethoids}
\resizebox{1.6 in}{!} {
\begin{tabular}{l|ccc|c}
\hline 
\hline 
Dataset & \multicolumn{3}{c|}{F1@\{10,25,50\}} & Acc \\
\hline 
Breakfast & $95.5$ & $87.5$ & $70.0$ & $76.9$\\
50Salads & $97.5$ & $90.4$ & $75.6$ & $81.3$\\
GTEA & $99.8$ & $97.7$ & $83.0$ & $75.3$\\
\hline 
\hline 
\end{tabular}
}
\end{table}

\begin{table}[h]
\centering
\caption{\textbf{Explicit duration prediction on Breakfast split 1.}
We show the results of different steps described in Section~\ref{sec:intro} from explicit duration prediction via the vanilla Transformer to implicit duration prediction with our alignment decoder and the impact of the full loss function, Eq.~\eqref{eq6}.}
\label{tab:expl_dur}
\resizebox{4.0 in}{!} {
\begin{tabular}{ll|ccc|c|c}
\hline 
\hline 
Method & Loss & \multicolumn{3}{c|}{F1@\{10,25,50\}} & Edit & Acc \\
\hline 
Vanilla Transformer, Fig.~\ref{fig:model_step_by_step}~(b) & Eq.~\eqref{eq:ce_segment} + MSE & $48.1$ & $42.3$ & $26.7$ & $52.9$ & $35.0$ \\
\hline
\multirow{2}{*}{+ Frame-wise Loss, Fig.~\ref{fig:model_step_by_step}~(c) } & Eq.~\eqref{eq:ce_segment} + Eq.~\eqref{eq:ce_frame} + MSE 
& $70.7$ & $63.5$ & $44.4$ & $73.9$ & $59.1$ \\
& Eq.~\eqref{eq6} + MSE & $72.1$ & $65.1$ & $48.7$ & $76.5$ & $59.0$ \\
\hline
\multirow{2}{*}{+ Alignment Decoder, Fig.~\ref{fig:model_step_by_step}~(d)+(e)} & Eq.~\eqref{eq:ce_segment} + Eq.~\eqref{eq:ce_frame} + AD & $73.5$ & $68.3$ & $54.3$ & $75.2$ & $67.7$ \\
& Eq.~\eqref{eq6} + AD & $77.1$ & $72.0$ & $60.4$ & $78.2$ & $71.7$ \\
\hline 
\hline 
\end{tabular}
}
\end{table}

\paragraph{Impact of the Loss Terms.}
In Table~\ref{tab:ablation_losses} we investigate the impact of the different loss terms (Section~\ref{sec:losses}) on split 1 of Breakfast and 50Salads.
In the first row of Table~\ref{tab:ablation_losses}, we evaluate the encoder when trained only using the frame-wise loss, \ie following the frame-wise prediction design as previous works. As expected, solely relying on the frame-wise loss leads to over-segmentation and poor performance. The rest of Table~\ref{tab:ablation_losses} shows the performance of our proposed model when using both encoder and decoder as explained in Sections~\ref{sec:model1} and \ref{sec:losses}, and reflect the key idea of our method to directly predict the segments.
While the most basic version using the segment-wise loss~\eqref{eq:ce_segment} improves over frame-wise predictions, we observe that using both the frame-wise~\eqref{eq:ce_frame} and segment-wise~\eqref{eq:ce_segment} loss term increases the performance drastically.
Moreover, we observe that adding the cross-attention loss~\eqref{eq:ca} further improves the results, demonstrating its effectiveness for longer sequences with many action segments, such as 50Salads.
While adding the group-wise loss terms \eqref{eq:g_frame} and \eqref{eq:g_segment} individually improves the performance moderately, the real benefit is revealed when combining them all together.

\begin{table}[h]
\centering
\caption{\textbf{Loss terms.} Contribution of different loss terms on Breakfast and 50Salads (split 1).}
\resizebox{4.0 in}{!} {
\begin{tabular}{l||ccc|c||ccc|c}
\hline 
\hline 
\multicolumn{1}{c||}{} & \multicolumn{4}{c||}{Breakfast} & \multicolumn{4}{c}{50Salads}\\
\cline{2-9} 
 & \multicolumn{3}{c|}{F1@\{10,25,50\}} & Edit & \multicolumn{3}{c|}{F1@\{10,25,50\}} & Edit \\
\hline 
$\mathcal{L}_{\text{frame}}$ & $8.9$ & $7.7$ & $5.9$ & $14.1$ & $13.5$ & $12.8$ & $10.8$ & $11.4$\\
$\mathcal{L}_{\text{segment}}$ & $49.5$ & $39.7$ & $22.9$ & $55.6$ & $20.1$ & $16.3$ & $8.6$ & $29.2$\\
$\mathcal{L}_{\text{frame}}$+$\mathcal{L}_{\text{segment}}$ & $71.8$ & $66.3$ & $52.6$ & $73.4$ & $55.0$ & $52.4$ & $37.5$ & $45.3$\\
$\mathcal{L}_{\text{frame}}$+$\mathcal{L}_{\text{segment}}$+$\mathcal{L}_{\text{CA}}$ & $73.8$ & $67.0$ & $54.8$ & $74.5$ & $74.2$ & $71.0$ & $58.4$ & $65.5$\\
$\mathcal{L}_{\text{frame}}$+$\mathcal{L}_{\text{segment}}$+$\mathcal{L}_{\text{g-frame}}$ & $73.3$ & $65.8$ & $52.8$ & $73.6$ & $56.6$ & $53.4$ & $40.2$ & $44.0$\\
$\mathcal{L}_{\text{frame}}$+$\mathcal{L}_{\text{segment}}$+$\mathcal{L}_{\text{g-segment}}$ & $72.8$ & $64.3$ & $53.7$ & $73.2$ & $59.1$ & $56.1$ & $42.8$ & $51.6$\\
$\mathcal{L}_{\text{frame}}$+$\mathcal{L}_{\text{segment}}$+$\mathcal{L}_{\text{g-frame}}$+$\mathcal{L}_{\text{g-segment}}$ & $73.5$ & $67.9$ & $55.0$ & $73.1$ & $57.0$ & $54.5$ & $40.4$ & $42.4$\\
$\mathcal{L}_{\text{frame}}$+$\mathcal{L}_{\text{segment}}$+$\mathcal{L}_{\text{g-frame}}$+$\mathcal{L}_{\text{g-segment}}$+$\mathcal{L}_{\text{CA}}$ & $75.1$ & $68.9$ & $54.9$ & $76.1$ & $73.6$ & $71.5$ & $55.3$ & $78.4$\\
\hline 
\hline 
\end{tabular}
\label{tab:ablation_losses}
}
\end{table}

\begin{figure}[h]
    \centering
    \includegraphics [scale=.4]{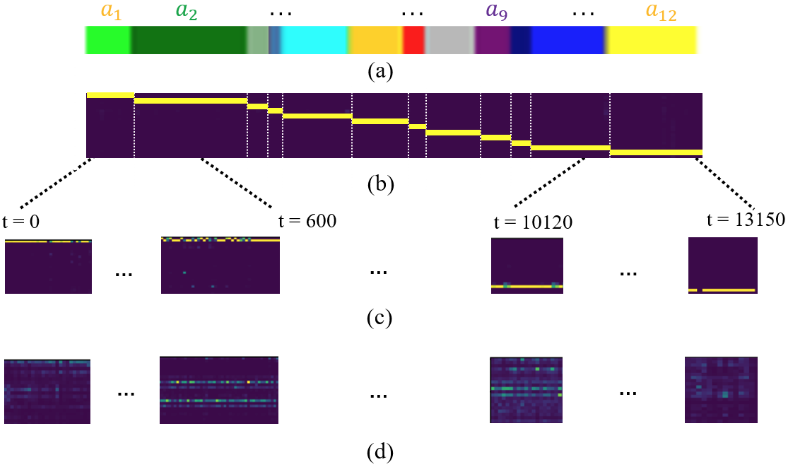}
    \caption{\textbf{Impact of the Cross-Attention Loss for the Transcript Decoder.} (a) A ground truth example of  a video with $13150$ frames and $12$ segments from the 50Salads dataset. (b) The target cross-attention map after softmax with dimension $12\times 13150$. (c) and (d) show the zoomed-in segments of the cross-attention map of the decoder when using the cross-attention loss (top) or not using it (bottom). 
    In (b-d) brighter color means higher values of the activations.}
    \label{fig:ca_ablation}
\end{figure}

To shed more light on the contribution of our cross-attention loss we visualize its impact in Fig.~\ref{fig:ca_ablation}.  
Given the ground truth segmentation, Fig.~\ref{fig:ca_ablation}~(a), of a video, Fig.~\ref{fig:ca_ablation}~(b) shows our expected target activations (output of softmax) of the decoder's cross-attention map; we hypothesize that activations should be higher in areas that belong to the corresponding segment.
Fig.~\ref{fig:ca_ablation}~(c) shows the output of the cross-attention when using our cross-attention loss. 
We observe that this loss indeed guides the cross-attention to have higher activations in the regions that belong to the related segment for an action.
Fig.~\ref{fig:ca_ablation}~(d) shows that lack of our cross-attention loss causes the attention map to be noisy; it's unclear which region is used for the segment classification.

\section{Conclusion}
We presented \textbf{\textit{UVAST}}, a new unified design for fully and timestamp supervised temporal action segmentation via Transformers in a seq2seq style. While the segment-level predictions of our model effectively address the over-segmentation problem, this new design entails a new challenge: predicting the duration of segments explicitly does not work out of the box. Therefore, we proposed three different approaches to alleviate this problem, enabling our model to achieve competitive performance on all three datasets.

{\footnotesize \textbf{Acknowledgements.} JG has been supported by the Deutsche Forschungsgemeinschaft (DFG, German Research Foundation) GA1927/4-2 (FOR 2535 Anticipating Human Behavior) and the ERC Starting Grant ARCA (677650).}

%
%

\clearpage
\bibliographystyle{splncs04}
\bibliography{main}

\setcounter{section}{0}
\section*{Unified Fully and Timestamp Supervised Temporal Action Segmentation via Sequence to Sequence Translation \\
Supplementary Material}

The structure of this supplementary material is as follows.
In Section~1, we provide more details regarding the datasets that we are using as well as highlighting the main differences among them in terms of number of videos, lengths, and segments.
In Section~2, we provide the implementation details of our architecture; we investigate the impact of the encoder architecture, our Split-Segment approach, and the constrained K-Medoids algorithm. Furthermore, we provide the values for our hyper-parameters. 
Last but not least, in Section~3, we provide more insight on our proposed grouping losses.

\section{Datasets}
In Table~\ref{tab:dataset_statistics}, we provide general information regarding the datasets (GTEA, 50Salads, and Breakfast) that we use for our experiments.
Among these datasets Breakfast has the largest number of classes and videos. On the other hand, the GTEA and 50Salads dataset are much smaller in terms of number of samples. The 50Salads dataset has the longest video sequences among the three datasets, while GTEA contains videos with the highest number of action segments and repetitions of an action within a video. 

\begin{table}[h!]
\centering
\caption{\textbf{Statistics of Action Segmentation Datasets.}}
\resizebox{\textwidth}{!} {
\begin{tabular}{c|c|c|c|c|c|c|c}
\hline 
\hline 
\multirow{2}{*}{Dataset} & \multirow{2}{*}{\# of Classes} & \multirow{2}{*}{\# of Videos} & \multirow{2}{*}{Min Videos Length} & \multirow{2}{*}{Max Videos Length} & \multirow{2}{*}{Mean Videos Length} & Min \# of Segments & Max \# of Segments\tabularnewline
 &  &  &  &  &  & in a Video & in a Video\tabularnewline
\hline 
GTEA \cite{fathi2011learning} & $11$ & $28$ & $634$  & $2009$ & $1115.18$ & $21$ & $44$\tabularnewline
50Salads \cite{stein2013combining} & $17$ & $50$ & $7555$ & $18143$ & $11551.90$ & $15$ & $26$\tabularnewline
Breakfast \cite{kuehne2014language} & $48$ & $1712$ & $130$ & $9741$ & $2097$ & $2$ & $25$\tabularnewline
\hline 
\hline 
\end{tabular}
}
\label{tab:dataset_statistics}
\end{table}

\section{Implementation Details}

All of the training and testing experiments were conducted on a single NVIDIA V100 GPU.


\subsubsection*{Impact of Encoder Model.}
\begin{figure}[h]
    \begin{center}
    \includegraphics[width=0.5\linewidth]{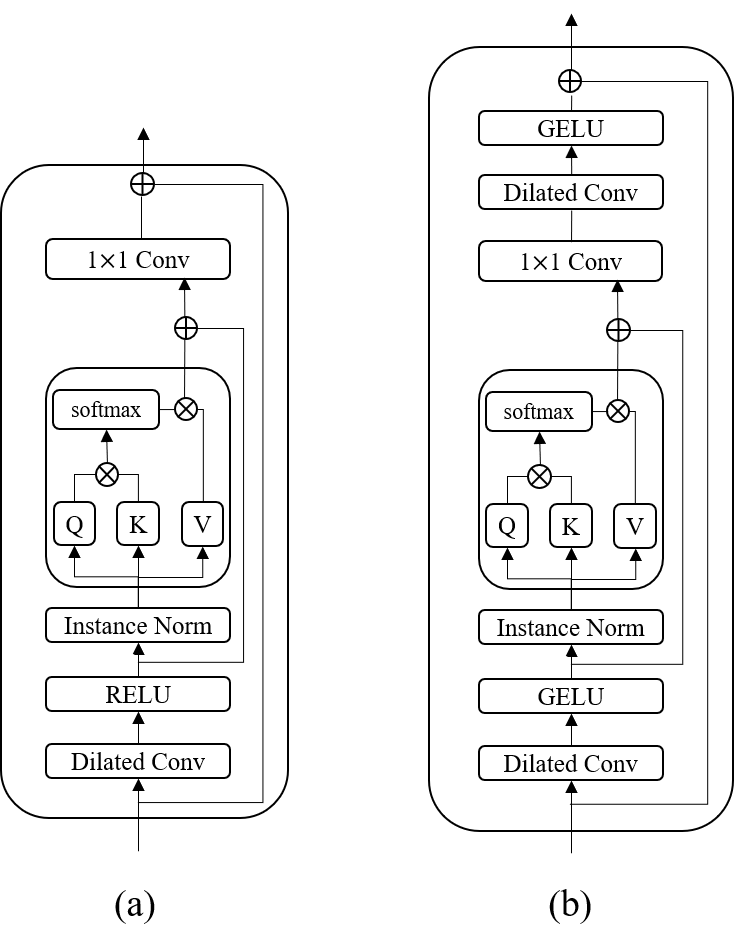}
    \end{center}
    \caption{\textbf{Modifications on Encoder Model.}
    Comparison between the original encoder block proposed by \cite{asformer} (a) 
    and our modified version (b) with \texttt{GELU} activations instead of \texttt{RELU} and an additional dilated convolution at the end of the encoder block.
    }
    \label{fig:enc}
\end{figure}

In our proposed algorithm, we utilized a modified version of the encoder model proposed in \cite{asformer}. The encoder model proposed by \cite{asformer} takes advantage of window attention as well as hierarchical representation for the
action segmentation task. While we find their proposed encoder model very effective and inspiring, we made small modifications to the architecture that further improved the performance.
Particularly, we replace the $\texttt{RELU}$ activation layers with $\texttt{GELU}$ activation and add one more layer of dilated convolution at the end of each encoder block. 
Fig.~\ref{fig:enc} shows the side by side comparison between the encoder block proposed by \cite{asformer} and our modified version.
Moreover, Table~\ref{enc_table} shows the Edit performance for using a simple encoder block (default pytorch implementation), the one proposed by \cite{asformer}, and our proposed modified version (last row in Table~\ref{enc_table}) on split 1 of GTEA, 50Salad, and Breakfast. 
While the ASFormer encoder achieves drastic improvements over the simple encoder, our modified version further improves over the ASFormer encoder.

\begin{table}[h!]
    \centering
    \caption{\textbf{Impact of Encoder Model.} Quantitative comparison (Edit score) between the impact of different encoder blocks on split 1 of GTEA, 50Salads, and Breakfast.}
    \resizebox{2.7 in}{!} {
    \begin{tabular}{ll|c|c|c}
    \hline 
    \hline 
     &  & Breakfast & 50Salads & GTEA\tabularnewline
    \hline
    \multirow{3}{*}{UVAST with} & Simple Encoder & $70.3$ & $68.7$ & $69.4$\tabularnewline
     & ASFormer \cite{asformer} Encoder & $74.6$ & $73.8$ & $88.6$\tabularnewline
     & Proposed Encoder & $76.1$ & $75.4$ & $93.4$\tabularnewline
    \hline 
    \hline 
    \multirow{2}{*}{UVAST Timestamp}
    & ASFormer \cite{asformer} Encoder & $72.9$ & $76.6$ & $89.0$ \tabularnewline
    & Proposed Encoder & $74.3$ & $78.3$ & $89.1$ \tabularnewline
    \hline
    \hline
    \end{tabular}
    }
    \label{enc_table}
\end{table}

\paragraph{Impact of Split-Segment.}
Due to a strong imbalance in the duration of action segments, we propose a \textit{split-segment} approach for improving the training of the network, where longer action segments are split up into several shorter ones, so that segment durations are more uniformly distributed; it is important to note that during the inference we \emph{do not} use any split-segment and use the video as is. In the timestamp supervised setting, we do not use split-segment as we do not have access to the ground truth duration of segments.
For the split-segment approach, we scale the durations to $[0, 1]$ by dividing the absolute duration of a segment (\textit{i.e.,} the number of frames in the segment) by the total number of frames in the video. 
For instance, if the split-segment value is set to $0.1$ it means that the duration of each action segment should be at most $0.1$ and segments larger than $0.1$  will be split up into smaller segments with maximum length of $0.1$.
During inference we merge the repeated actions into one: For example, if our model predicts an action sequence of $(A,B,B,C,A,A,A)$ we convert it to $(A,B,C,A)$.
The split-segment value is a hyper parameter that can be selected empirically for each dataset.
In Table~\ref{split} we show the impact of using split-segment on split 1 of the Breakfast, 50Salads, and GTEA datasets, where we report the Edit scores. We can see that using the split-segment approach helps the model achieve better performances on all three datasets.
Furthermore, in Table~\ref{split_values} we provide an ablation study regarding the impact of selecting different split-segment values on the split 1 of the Breakfast dataset.

\begin{table}[h!]
\parbox{.48\linewidth}{
\centering
\caption{\textbf{Impact of Split-Segment.} Quantitative comparison (Edit score) between the impact of using split-segment versus not using it on split 1 of GTEA, 50Salads, and Breakfast dataset.}
\resizebox{.9\linewidth}{!} {
\begin{tabular}{l|c|c|c}
\hline 
\hline 
 & Breakfast & 50Salads & GTEA\tabularnewline
\hline 
No Split-Segment & $75.0$ & $74.2$ & $88.2$\tabularnewline
With Split-Segment & $76.1$ & $75.4$ & $93.4$\tabularnewline
\hline 
\hline 
\end{tabular}}
\label{split}
}
\hfill
\parbox{.48\linewidth}{
\centering
\caption{\textbf{Impact of Split-Segment Values.} Ablation study on split 1 of the Breakfast dataset regarding the impact of using different values for the split-segment on the performance (Edit score).}
\label{split_values}
\resizebox{.8\linewidth}{!} {
\begin{tabular}{l|cccccc}
\hline 
\hline 
\multirow{2}{*}{} & \multicolumn{6}{c}{Split-Segment Values}\tabularnewline
 & $0.05$ & $0.1$ & $0.15$ & $0.17$ & $0.2$ & $0.3$\tabularnewline
\hline 
Breakfast & $74.9$ & $75.6$ & $76.1$ & $76.1$ & $76.1$ & $75.4$\tabularnewline
\hline 
\hline 
\end{tabular}
}
\label{split2}
}
\end{table}

\paragraph{Hyper-Parameters.}
Table~\ref{param} provides a summary of the values for our hyper-parameters used for training our model.
In Table~\ref{param}, Stage 1 and 2 refer to Section 3.1 and 3.3 of the main paper, where we train the  Transformer for auto-regressive segment prediction and alignment Transformer for duration prediction, respectively. Furthermore, as shown in Table~\ref{param}, we use a cross-attention smoothing (average pooling on the cross-attention map along the $T$ dimension) for the 50Salads dataset to reduce the noise in the cross-attention. Comparing to the Breakfast and GTEA datasets, 50Salads dataset has the longest videos (see Table~\ref{tab:dataset_statistics}) with a very low number of training data which causes the cross-attention map to be noisy. Our proposed cross-attention loss along with cross-attention smoothing  helps to reduce the noise and therefore leads to a better performance.
Similar to previous methods and to have a fair comparison, we use sampling rate of $2$ for the 50Salads dataset since it has higher FPS compared with the other two datasets \cite{li2020ms}.
We state the hyper-parameters used in FIFA and Viterbi in Table~\ref{tab:hparam}.

\begin{table}[h!]
\centering
\caption{\textbf{Hyper-Parameters.}}
\resizebox{\linewidth}{!} {
\begin{tabular}{c|c|c|c|c|c|c|c|c|c|c|c|c|c|c|c|c|c|c|c|c}
\hline 
\hline 
 & \multicolumn{9}{c|}{Common Between Stage 1 and 2} & \multicolumn{5}{c}{Stage 1 (Encoder-Decoder)} &  & \multicolumn{3}{c}{Stage 2 (Alignment Decoder)}\tabularnewline
\hline 
 & Batch & Optimizer & LR & Epoch & sampling rate & d & d' & $\tau'$ & Dropout & Activation & Split Segment & \# Layers in & \# Layers in & Decoder Feedforward & Cross-Attention & Smoothing & \# Parameters & \# Layers in & Decoder Feedforward & \# Parameters\tabularnewline
 & Size &  &  &  &  &  &  &  &  &  &  & Encodeer & Decoder & Dimension & Smoothing & Kernel &  & Alignment Decoder & Dimension & \tabularnewline
\hline 
Breakfast & 1 & adam & 0.0005 & 800 & 1 & 2048 & 64 &0.001& $\checkmark$ & GELU & 0.17 & 10 & 2 & 2048 & $\times$ & $\times$ & 1.109M & 1 & 1024 & 0.166M\tabularnewline
50Salads & 1 & adam & 0.0005 & 800 & 2 & 2048 & 64 &0.001& $\checkmark$ & GELU & 0.15 & 10 & 2 & 2048 & $\checkmark$ & 31 & 1.103M & 1 & 1024 & 0.166M\tabularnewline
GTEA & 1 & adam & 0.0005 & 800 & 1 & 2048 & 64 &0.001& $\checkmark$ & GELU & 0.17 & 10 & 2 & 2048 & $\times$ & $\times$ & 1.102M & 1 & 1024 & 0.166M\tabularnewline
\hline 
\hline 
\end{tabular}
}
\label{param}
\end{table}

\begin{table}[h!]
\centering
\caption{\textbf{Hyper-Parameters used in FIFA and Viterbi.}}
\label{tab:hparam}
\resizebox{0.4\linewidth}{!} {
\begin{tabular}{l|ccc|c}
\hline 
\hline 
 Dataset & \multicolumn{3}{c|}{FIFA} & Viterbi \tabularnewline
 & Epochs & Sharpness & Step-size & frame sampling\tabularnewline
\hline 
Breakfast & $3000$ & $80$ & $0.01$ & $5$ \tabularnewline
50Salads & $3000$ & $80$ & $0.01$ & $2$ \tabularnewline
GTEA & $3000$ & $80$ & $0.1$ & $1$ \tabularnewline
\hline 
\hline 
\end{tabular}}
\end{table}

\paragraph{K-Medoids.}
In the main paper, we propose a constrained k-medoids algorithm for generating pseudo-segmentations given frame-wise input features and timestamps. In contrast to the vanilla k-medoids clustering algorithm, our constrained version ensures temporal consistency of the clusters and the resulting temporally continuous clusters can be unambiguously identified with the class labels of the ground truth transcript. This is a major advantage over an unconstrained clustering method, which may result in temporally fragmented clusters, making class label assignment ambiguous. 
We compare our constrained k-medoids with the unconstrained version, see Table~\ref{tab:kmedoids_constrained_vs_unconstrained}, where we assign each cluster the class label belonging to the timestamp it was initialized with. Note, that in this scenario the original ground truth timestamps may end up in completely different clusters and the class label assignment becomes noisy.

\begin{figure}[h!]
\centering
\includegraphics[width=.9\linewidth]{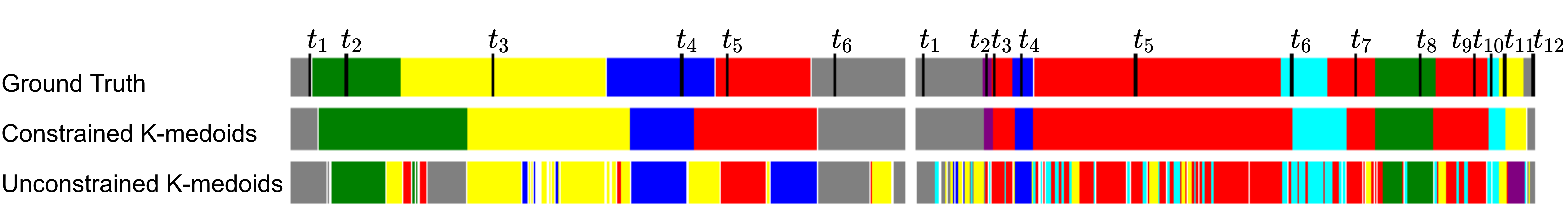}
\caption{\textbf{K-Medoids.}
We compare our constrained and the unconstrained k-medoids algorithm qualitatively. Both algorithms cluster the frame-wise input features, using the ground truth timestamps $t_1,\dots,t_n$ as initialization.
}
\label{fig:kmedoids_segmentation}
\end{figure}

As expected the temporal fragmentation of the clusters leads to over-segmentation and correspondingly low Edit and F1 scores. However, even on Acc this unconstrained version suffers due to the noisy label assignment.
Furthermore, we show two example videos from the Breakfast dataset in Fig.~\ref{fig:kmedoids_segmentation}; again, we observe over-segmentation due to temporally fragmented clusters.
Notably, we observe that the unconstrained k-medoids performs much better on GTEA compared with Breakfast and 50Salads. One reason for this can be the frequency of background classes, which are visually distinct to the action classes in the video and typically show very static scenes. The frequent background classes of highly similar features separate the action classes from one another. In contrast, Breakfast and 50Salads do not have a frequent background class and relatively static scenes are assigned to an action class, making it more difficult to separate them.

\begin{table}
\centering
\caption{\textbf{K-Medoids.} We compare our constrained k-medoids algorithm proposed in the main paper with the vanilla unconstrained version.
}
\label{tab:kmedoids_constrained_vs_unconstrained}
\resizebox{4.1 in}{!} {
\begin{tabular}{l|ccc|c|c||ccc|c|c||ccc|c|c}
\hline 
\hline 
\multicolumn{1}{l|}{} & \multicolumn{5}{c||}{\textcolor{black}{Breakfast}} & \multicolumn{5}{c||}{\textcolor{black}{50Salads}} & \multicolumn{5}{c}{\textcolor{black}{GTEA}}\tabularnewline
\cline{2-16}  
 & \multicolumn{3}{c|}{\textcolor{black}{F1@}} &  &  & \multicolumn{3}{c|}{\textcolor{black}{F1@}} & \multicolumn{1}{c|}{} &  & \multicolumn{3}{c|}{\textcolor{black}{F1@}} & \multicolumn{1}{c|}{} & \tabularnewline
 & \textcolor{black}{\{10} & \textcolor{black}{25} & \multicolumn{1}{c|}{\textcolor{black}{50\}}} & \textcolor{black}{Edit} & \textcolor{black}{Acc} & \textcolor{black}{\{10} & \textcolor{black}{25} & \multicolumn{1}{c|}{\textcolor{black}{50\}}} & \multicolumn{1}{c|}{\textcolor{black}{Edit}} & \textcolor{black}{Acc} & \textcolor{black}{\{10} & \textcolor{black}{25} & \multicolumn{1}{c|}{\textcolor{black}{50\}}} & \multicolumn{1}{c|}{\textcolor{black}{Edit}} & \textcolor{black}{Acc}\tabularnewline
\hline 
Constrained k-medoids & $95.5$ & $87.5$ & $70.0$ & $100.0$ & $76.9$ & $97.5$ & $90.4$ & $75.6$ & $100.0$ & $81.3$ & $99.8$ & $97.7$ & $83.0$ & $100.0$ & $75.3$\tabularnewline
Unconstrained k-medoids & $8.4$ & $6.6$ & $3.8$ & $12.3$ & $53.8$ & $3.8$ & $2.5$ & $1.0$ & $2.3$ & $52.9$ & $71.0$ & $68.2$ & $52.9$ & $59.8$ & $69.6$\tabularnewline
\hline 
\hline 
\end{tabular}
}
\end{table}

\section{Grouping Loss Terms}

The action segmentation datasets mentioned above are highly imbalanced in terms of the frequency of the actions that appear in the videos and the number of frames each action occupies, which influences the cross-entropy loss on top of the encoder and decoder. To cope with the imbalanced classes, we utilize two modified versions of the cross-entropy loss. The first loss involves averaging the probabilities of each class separately,
and the second one involves averaging the logits of each class before passing them to \texttt{Softmax}.

To shed more light on the intuition behind the modifications, let us consider a classifier that classifies $N$ frames $x_n$ into $C$ classes. We denote the logits by $a_{n}$ and the corresponding probabilities by $\mu_{n}=\texttt{Softmax}(a_{n})$ where $\texttt{Softmax}(a_{n})_c=\frac{e^{a_{n,c}}}{\sum_{i=1}^C e^{a_{n,i}}}$.
For grouping the frames by class label, we define:
\begin{align}
    N_c &= \{n|y_n = c\}\quad \text{for } c\in\{1,\dots,C\},\\
    \bar \mu_c &= \frac{1}{|N_c|}\sum_{n \in N_c} \mu_n, \\
    \hat \mu_c &= \texttt{Softmax}(\hat{a})_c\quad \hat{a}_c=\frac{1}{|N_c|} \sum_{n \in N_c} a_{n,c}.
\end{align}
We consider the following three loss terms, where the first is an element-wise cross-entropy loss and the last two group-wise cross-entropy loss terms, taking the average outside and inside the \texttt{Softmax}, respectively:
\begin{align}
    \mathcal{L}^{}&=-\sum_{n=1}^{N}\sum_{c=1}^Cy_{n_c}\log\mu_{n_c}, \label{eq_L} \\
    \bar{\mathcal{L}}^{}&=-\sum_{c=1}^C\log\bar \mu_c, \label{eq_L_bar} \\
    \hat{\mathcal{L}}^{}&=-\sum_{c=1}^C\log\hat \mu_c. \label{eq_L_hat}
\end{align}

Table~\ref{tbl:table8} summarizes the results for different choices of the grouping loss. We observe that $\bar{\mathcal{L}}$ works the best for the segment-wise loss on top of the decoder, and $\hat{\mathcal{L}}$ works best for the frame-wise classification. Note that we report the results of the first stage training, \ie transcript prediction only and therefore only report Edit score.

\begin{table}
\centering
\caption{\textbf{Impact of different modifications on the group loss.} Ablation study on split 1 of the 50Salads dataset regarding the impact of using different modifications of the group loss. We report Edit results for stage 1 training.
}
\resizebox{1.52 in}{!} {
\begin{tabular}{l|l|c}
\hline 
\hline 
$\mathcal{L}_{\text{g-frame}}$ & $\mathcal{L}_{\text{g-segment}}$ & 50Salads\tabularnewline
\hline 
$\bar{\mathcal{L}}$, Eq.~\eqref{eq_L_bar} & $\bar{\mathcal{L}}$, Eq.~\eqref{eq_L_bar} & $78.4$\tabularnewline
$\hat{\mathcal{L}}$, Eq.~\eqref{eq_L_hat} & $\hat{\mathcal{L}}$, Eq.~\eqref{eq_L_hat} & $78.1$\tabularnewline
$\bar{\mathcal{L}}$, Eq.~\eqref{eq_L_bar} & $\hat{\mathcal{L}}$, Eq.~\eqref{eq_L_hat} & $79.8$\tabularnewline
\hline 
\hline 
\end{tabular}
\label{tbl:table8}
}
\end{table}

\end{document}